\documentclass[sigconf]{acmart}
\AtBeginDocument{%
  }

\setcopyright{acmlicensed}
\copyrightyear{2026}
\acmYear{2026}
\acmDOI{XXXXXXX.XXXXXXX}
\acmConference[Conference acronym 'XX]{Make sure to enter the correct
  conference title from your rights confirmation email}{June 03--05,
  2026}{Woodstock, NY}
\acmISBN{978-1-4503-XXXX-X/2026/06}

\begin{document}

\title{The Expense of Seeing: Attaining Trustworthy Multimodal Reasoning Within the Monolithic Paradigm}

\author{Karan Goyal}
\email{karang@iiitd.ac.in}
\affiliation{%
  \institution{IIIT Delhi, India}
}

\renewcommand{\shortauthors}{Karan Goyal}


\begin{abstract}
The rapid proliferation of Vision-Language Models (VLMs) is often framed as enabling unified multimodal knowledge discovery but rests on an under-examined assumption: that current VLMs faithfully synthesise multimodal data. We argue they often do not, and this gap reflects a trustworthiness problem in the dominant Vision Encoder-Projector-LLM paradigm. Rather than extracting grounded knowledge from visual inputs, state-of-the-art models frequently exhibit functional blindness, i.e., exploiting strong language priors to bypass severe visual representation bottlenecks. In this work, we challenge the conventional methodology of multimodal evaluation, which relies on data ablation or new dataset creation and therefore conflates dataset biases with architectural incapacity. We propose an information-theoretic departure: the \textit{Modality Translation Protocol}, designed to quantify what we call the \textit{Expense of Seeing}. By translating semantic payloads rather than ablating them, we formulate three novel metrics—the Toll ($ToS$), Curse ($CoS$), and Fallacy ($FoS$) of Seeing—culminating in the \textit{Semantic Sufficiency Criterion (SSC)}. Furthermore, we hypothesise a \textit{Divergence Law of Multimodal Scaling}: as the underlying language engines scale to unprecedented reasoning capabilities, the penalty of the visual knowledge bottleneck may increase rather than diminish. We argue the community should move beyond ``multimodal gain'' as a primary evaluation target. By elevating the SSC from a passive diagnostic constraint to an active architectural blueprint, we provide a foundation for guiding the next generation of AI systems toward genuine multimodal reasoning.
\end{abstract}

\ccsdesc[300]{Computing methodologies~Knowledge representation and reasoning}

\keywords{Trustworthy AI, Modern AI, Foundations of Knowledge Representation, Decoding Multimodal Decision-making}

\maketitle

\section{The Illusion of Multimodal Synthesis}

The trajectory of Knowledge Discovery and Data Mining (KDD) has reached a critical inflection point. We are no longer merely mining tabular databases, massive graphs or isolated text corpora; the frontier of Modern AI and Big Data is the construction of unified ``world models'' \cite{nvidia_world_models}. These systems are expected to ingest and seamlessly synthesise disparate, high-dimensional information from text, images, videos, and complex topological graphs etc. to perform faithful, cross-domain decision-making. At the core of this frontier sits the Vision-Language Model (VLM), predominantly governed by the monolithic Vision Encoder-Projector-LLM architectural paradigm \cite{apple_fastvlm_2025,ibm_vlm, nvidia_vlm,bordes2024introduction}. 

The prevailing assumption within the global AI community is that these models natively integrate visual and textual streams to execute Compositional Visual Reasoning (CVR) \cite{ke2025explain}. Yet, as VLMs are increasingly deployed in high-stakes Data Science applications ranging from autonomous medical diagnostics \cite{hartsock2024vision,zhong2025vision} to financial time-series forecasting \cite{khezresmaeilzadeh2025morfi}, a notable epistemic fragility has been documented. Highly parameterised, state-of-the-art models frequently achieve superficial benchmark supremacy by largely ignoring the visual input. Instead, they exhibit a modern \textit{Clever Hans} effect, executing complex statistical guessing via deeply ingrained text priors housed within their massive Large Language Model (LLM) backbones \cite{choi2024improving,10.5555/3737916.3738766,chen2026babyvision}. A latest work, BabyVision \cite{chen2026babyvision}, shows that SOTA VLMs consistently fail on basic visual tasks that humans, even 3-year-olds, can solve effortlessly. It released a dataset benchmark designed to assess core visual abilities independent of linguistic knowledge for VLMs. MMVP \cite{tong2401eyes}
released a dataset benchmark to probe visual limitations and found that models fail to distinguish images with clear perceptual differences. ConMe \cite{huang2024conme} released compositional reasoning benchmark to produce `hard CR Q\&A'.

Recent initiatives within the representation learning community, such as MATHVERSE \cite{zhang2024mathverse}, SeePHYS \cite{xiang2025seephys} and MMStar \cite{10.5555/3737916.3738766}, have attempted to uncover and quantify this phenomenon. MATHVERSE \cite{zhang2024mathverse} introduced problem versions with varying visual-textual information balance and observed that some VLMs achieve higher accuracy when visual input was removed entirely. SeePHYS \cite{xiang2025seephys} extended this to Physics, distinguishing ``vision-essential” from ``vision-optional” problems. MMStar \cite{10.5555/3737916.3738766} proposed heuristic metrics like Multimodal Gain and Multimodal Leakage, and a manually vetted vision-indispensable dataset for multimodal assessment. However, we assert that these approaches violate the rigorous foundations of knowledge discovery. By ablating (removing) data to test models, they successfully expose \textit{dataset biases} but fundamentally fail to isolate \textit{architectural representation bottlenecks}. We cannot map the limits of a model's knowledge extraction prowess by measuring what happens when knowledge is artificially deleted.

In an era where modern reasoning engines achieve increasingly strong symbolic logic, a difficult question arises: \textit{\textbf{What if vision is no longer a value-add for knowledge discovery in its present form, but an active architectural liability?}} This paper introduces a framework to systematically diagnose, quantify, and address these integration failures. We shift the paradigm from observing macroscopic, dataset-induced heuristics to establishing principled, sample-level diagnostic criteria for Trustworthy and Responsible Data Science.\textbf{ We propose the field shift from measuring additive Multimodal Gain or creating new dataset benchmarks towards diagnosing the \textit{Expense of Seeing}} ---  a step toward constructing genuinely faithful world models within the monolithic paradigm.

\section{Challenging Existing Assumptions: The Crisis in Evaluation}

To build trustworthy data science systems, our evaluation metrics must rigorously isolate the \textit{source} of a model's predictive power. The standard approach of evaluating multimodal capabilities currently relies heavily on the paradigm of data ablation. 

\subsection{The Flaws of Multimodal Gain and Leakage}
Consider the formulation of Multimodal Gain ($MG$), which measures the difference in accuracy when a model is given both vision and text ($S_v$) versus text alone ($S_{wv}$): $MG = S_v - S_{wv}$. Similarly, Multimodal Leakage ($ML$) assesses leakage by comparing the VLM’s text-only performance against its underlying base LLM ($S_t$): $ML = \max(0, S_{wv} - S_t)$.

We assert that these metrics are limited for evaluating modern world models. 
\begin{enumerate}
    \item \textbf{Biased Estimators:} $ML$ is limited as a global estimator. By utilising a $\max$ function, it routinely fails to account for destructive interference scenarios where the multimodal alignment training process degrades the base LLM’s inherent reasoning capabilities ($S_{wv} < S_t$). 
    \item \textbf{The Ablation Fallacy:} $MG$ does not measure faithful integration; it measures the leverage of an additional signal under conditions of artificial starvation. From an information-theoretic perspective, if we deprive a model of required information (by deleting the image), any subsequent failure cannot be definitively attributed to an architectural inability to process vision. Measuring what happens when information is removed does not directly characterise a model's capacity to extract information when it is present.
\end{enumerate}

\subsection{The Necessity of a Paradigm Shift}
To build trustworthy systems capable of synthesising powerful combinations of visual and textual data, we must move beyond data ablation and the race to new dataset creation. We must isolate the \textit{architectural bottleneck} from the \textit{dataset bias}. If the research community continues to use ablative metrics, we risk deploying models that rely on language priors rather than grounded visual evidence, leading to high-cost failures in real-world applications.

\subsection{Operationalising the Paradigm Shift: A New Diagnostic Agenda}
To successfully transition from observing dataset-induced heuristics to diagnosing fundamental architectural bottlenecks, the KDD community must align around a new empirical standard. Based on the necessity of preserving semantic equivalence, we propose that the future evaluation of multimodal world models must be anchored by six operational, highly testable research questions:

\begin{itemize}
    \item \textbf{[RQ1] The Baseline Penalty:} Do current VLM architectures incur a systematic, quantifiable performance penalty when extracting knowledge from visual inputs compared to processing equivalent (and potentially even lossy) symbolic textual representations?
    
    \item \textbf{[RQ2] The Architectural Origin:} Can we mathematically distinguish and isolate whether a model's inefficiency originates in the visual encoder (an incapacity to read visual features) or the cross-modal projection head (an incapacity to fuse separate semantic streams)?
    
    \item \textbf{[RQ3] Semantic Asymmetry:} Do VLMs exhibit semantic inconsistency across modalities? If provided with equivalent information in symbolic textual and symbolic visual forms, does the architecture asymmetrically penalise the act of ``seeing'' rather than ``reading''?
    
    \item \textbf{[RQ4] The Scaling Paradox:} Within a fixed architectural family, what does drastically increasing parameter scale actually achieve? Does scaling the underlying language engine alleviate the visual bottleneck, or paradoxically exacerbate it?
    
    \item \textbf{[RQ5] The Universal Constraint:} Can we design a singular, mathematically sound criterion that detects, quantifies, and localises these multimodal failures across any given architecture?
    
    \item \textbf{[RQ6] Dataset Agnosticism:} Can a diagnostic toolkit definitively prove that an integration failure is caused by an architectural bottleneck rather than dataset bias, eliminating the field's reliance on data ablation and specially vetted ``vision-indispensable'' benchmarks?
\end{itemize}

\section{The Modality Translation Protocol \& High-Stakes KDD Case Studies}

We propose a new methodological approach: The Modality Translation Protocol. Instead of deleting information to test a model, this protocol preserves the exact semantic payload of a data sample while translating its modality across different representation states. 

Let $S(\cdot)$ denote the primary evaluation metric (e.g., Accuracy, Exact Match) of a model $\mathcal{M}$ on a given task. For any single multimodal data sample, we define three distinct modulations:

\begin{enumerate}
    \item \textbf{$S_{Full}$ (Standard VLM):} Evaluated with standard visual input $V$ and textual input $T$.
    \begin{equation}
        S_{Full} = S(\mathcal{M}(V, T))
    \end{equation}
    
    \item \textbf{$S_{SymT}$ (Symbolic Text Ceiling):} The visual input $V$ is replaced by a task-sufficient symbolic text representation $V_{label}$. This measures the LLM's task-relevant reasoning capacity given symbolic access to the same evidence the visual modality is meant to convey.
    \begin{equation}
        S_{SymT} = S(\mathcal{M}(\emptyset, T + V_{label}))
    \end{equation}
    
    
    \item \textbf{$S_{SymV}$ (Symbolic Vision):} The textual question $T$ is rendered perfectly as text-within-an-image $T_{img}$, forcing the model to read solely via its visual encoding pipeline without discrete text tokens.
    \begin{equation}
        S_{SymV} = S(\mathcal{M}(V + T_{img}, \emptyset))
    \end{equation}
\end{enumerate}

We illustrate how this protocol exposes architectural failures across three critical domains of Knowledge Discovery.

\begin{itemize}
    \item \textbf{Case Study 1: Financial Time-Series Mining:} A VLM analyses a candlestick chart to predict a breakout. $S_{SymT}$ replaces the chart with perfect OHLC tabular text. If $S_{SymT}$ yields $95\%$ accuracy but $S_{Full}$ yields $60\%$, the underlying LLM demonstrates competence at financial reasoning, but the visual encoder actively bottlenecks knowledge extraction.
    \item \textbf{Case Study 2: Trustworthy Medical Diagnostics:} A VLM evaluates a chest X-Ray, prompted with clinical notes: \textit{``Patient has a 30-year history of smoking.''} Driven by text priors, $S_{Full}$ predicts cancer. $S_{SymT}$ replaces the image with ground-truth symbolic findings: \textit{``Clear lungs.''} If $S_{SymT}$ correctly predicts ``Healthy'' but $S_{Full}$ hallucinates cancer, we expose a serious text-prior override.
    \item \textbf{Case Study 3: Molecular Graph Mining for Drug Discovery:} A VLM screens a 2D molecular structure for toxicity based on visual topology and textual properties. $S_{SymV}$ removes the text prompt, rendering the text directly into the 2D molecule image. If $S_{Full}$ underperforms $S_{SymV}$, the projection head cannot align continuous visual coordinate spaces with discrete token spaces. And if it outperforms, then it indicates inefficiency in visual encoding.
\end{itemize}

\subsection{Constructing \texorpdfstring{$V_{label}$}{Vlabel}: Task-Conditional Sufficiency}
The protocol's diagnostic power rests on a precise construction of $V_{label}$. We do not require $V_{label}$ to be a strictly lossless information-theoretic translation of $V$, which is generally unachievable: a chest X-ray contains continuous pixel-level information that no finite symbolic description fully captures. Instead, we define $V_{label}$ as \textit{task-sufficient}: it preserves all task-relevant discriminative information that an idealised observer could extract from $V$ to answer the task. Under this definition, $S_{SymT}$ characterises the LLM's reasoning capacity on the task given symbolic access to the same evidence, and as detailed in Section \ref{ToS section},  $ToS > 0$ admits a clean interpretation: \textit{given equivalent task-relevant information, the model performs worse when that information is delivered visually than symbolically.}

This formalisation also clarifies how $V_{label}$ should be constructed in practice. In a large and important class of KDD tasks, the visual modality is itself a rendering of underlying structured data; here $V_{label}$ recovers the pre-rendering symbolic form. The three case studies above instantiate this pattern: OHLC tabular text underlies the candlestick chart; ground-truth radiological findings underlie the X-ray label; SMILES strings or atom-bond lists underlie the molecular diagram. Where no pre-rendering form exists, $V_{label}$ is constructed via oracle annotation against the task's ground truth (expert annotation, structured database lookup, or symbolic extraction pipelines).

This scoping is a feature, not a concession. The framework applies cleanly to the broad class of tasks central to scientific data mining, structured visual reasoning, chart and diagram understanding, annotated medical imaging, molecular and graph-based discovery, and document intelligence --- precisely the high-stakes settings where trustworthy multimodal evaluation matters most. Open-ended perceptual tasks (e.g., describing the mood of a photograph) fall outside this scope by design, and we make no claim about them.

\section{True Quantifiers of Visual Reception}

Utilising the Modality Translation Protocol, we define three novel metrics that act as quantifiable indicators of multimodal knowledge bottlenecks. These metrics move the community beyond asking \textit{whether} a model works, to diagnosing \textit{why, how much, and where} multimodal reasoning breaks down.

\subsection{\texorpdfstring{$ToS$}{ToS}: Toll of Seeing}
\label{ToS section}
The actual expense the VLM bears to process the visual modality, operationalising the baseline penalty of integration:
\begin{equation}
    ToS = S_{SymT} - S_{Full}
\end{equation}
\textbf{Diagnostic Interpretation [RQ1]:} Ideally, $ToS \le 0$. If $ToS > 0$, we identify an architectural inefficiency in visual encoding and/or integration. The model incurs a systematic performance penalty when processing visual input compared to its equivalent (and potentially even lossy) symbolic textual representation. Vision acts as a toll on the LLM's inherent reasoning capacity.

\subsection{\texorpdfstring{$CoS$}{CoS}: Curse of Seeing}
The asymmetric penalty of processing information across different modalities:
\begin{equation}
    CoS = S_{SymT} - S_{SymV}
\end{equation}
\textbf{Diagnostic Interpretation [RQ3]:} Ideally, $CoS \le 0$. If $CoS > 0$, the architecture exhibits semantic inconsistency. It reveals an asymmetric penalisation of \textit{seeing} rather than \textit{reading} equivalent information (and potentially even lossy). A faithful multimodal model should treat semantically equivalent inputs symmetrically; $CoS > 0$ indicates the model is systematically biased against non-textual knowledge extraction.

\subsection{\texorpdfstring{$FoS$}{FoS}: Fallacy of Seeing}
The third metric of our diagnostic resolution, distinguishing the exact origin of the architectural bottleneck:
\begin{equation}
    FoS = S_{Full} - S_{SymV}
\end{equation}
The asymmetry between $S_{Full}$ (dual-stream) and $S_{SymV}$ (single-stream) is intentional: it is precisely this contrast that enables the sign of $FoS$ to localise failure to encoder versus projector, since the two conditions hold the semantic payload constant while varying how the model must fuse it.

\textbf{Diagnostic Interpretation [RQ2]:} Mathematically, $FoS \equiv CoS-ToS$. However, we must explicitly define and evaluate $FoS$ separately because it diagnoses a distinct failure mode. Ideally, $FoS = 0$. Humans process $S_{Full}$ and $S_{SymV}$ equally well without loss of fidelity. If $FoS \neq 0$, it is a fallacy that this exists and confirms the architecture's inability to process the same lossless semantic payload consistently across these conditions. Crucially, the sign of $FoS$ reveals two distinct, mutually exclusive failure modes:
\begin{itemize}
    \item \textbf{The Positive Collapse Mode ($FoS > 0$):} Indicates an inefficiency in visual encoding. The model struggles to read and extract text when it is rendered purely as an image, indicating the vision encoder (e.g., the ViT) lacks the granular spatial resolution to extract symbolic features.
    \item \textbf{The Negative Collapse Mode ($FoS < 0$):} Indicates an inefficiency in visual integration. The model performs paradoxically better when forced into a single visual modality ($S_{SymV}$) than when handling separate visual and textual streams ($S_{Full}$). This isolates the failure to the cross-modal projection head, indicating it cannot meaningfully fuse separate modalities in the latent space.
\end{itemize}

\section{The Semantic Sufficiency Criterion (SSC)}

Together, these metrics establish a mandatory mathematical condition for semantically grounded, faithful multimodal data science. We define the Semantic Sufficiency Criterion (SSC):
\begin{equation}
    SSC : \max(ToS, CoS, |FoS|) = 0
\end{equation}

\subsection{A Diagnostic Constraint, Not an Immediate Goal}
Crucially, the SSC is not treated as an immediately achievable performance target for current models, but as a stringent diagnostic constraint \textbf{[RQ5]}. Violations of the SSC (where $SSC > 0$) quantify the exact magnitude and location of a VLM's failure. The application of the absolute value $|FoS|$ is necessary to ensure that both the positive (encoding) and negative (integration) failure modes are captured and audited. 

\subsection{The KDD Advantage: Universal Dataset Applicability}
Because the protocol never ablates either signal (in contrast to MMStar \cite{10.5555/3737916.3738766}), failures detected by SSC can be attributed to \textit{architectural bottlenecks} rather than dataset-induced artifacts. Consequently, this diagnostic toolkit provides a great advantage for KDD researchers: it can be applied to any regular dataset \textbf{[RQ6]}. We no longer need to rely on specially vetted datasets to test models. The SSC identifies and quantifies architectural violations across multimodal tasks.

\section{The Divergence Law of Multimodal Scaling}
The dominant orthodoxy in \textit{Systems for Data Science and Scalable AI} dictates a simple heuristic: scaling the compute and parameter counts inherently resolves multimodal alignment challenges. The industry operates on the foundational assumption that concatenating progressively larger Vision Transformers (ViTs) to progressively larger Large Language Models (LLMs) will organically yield faithful multimodal synthesis. \textbf{We argue this assumption is unlikely to hold.} 

By operationalising our diagnostic agenda, specifically addressing \textbf{[RQ4]}, we posit that the current architectural paradigm is structurally limited in achieving true multimodal knowledge discovery. Scale-driven alignment may mask, rather than resolve, a structural failure.

\subsection{The Mechanics of the Representation Bottleneck}
\begin{sloppypar}
To understand why scaling fails, we must examine the information-theoretic capacity mismatch inherent in the Vision Encoder-Projector-LLM paradigm. Visual data manifolds (e.g., the topology of a molecular graph or the high-frequency pixel variations in a medical scan) are continuous, high-dimensional, and dense. Conversely, textual token spaces are discrete, sequential, and highly compressed.
\end{sloppypar}

Current architectures force the entirety of the visual manifold to pass through a narrow, fixed-capacity cross-attention or projection head to be translated into text-like embeddings. As the LLM backbone scales, its capacity to execute complex symbolic logic and leverage statistical priors outpaces the projection head’s ability to faithfully translate the visual manifold. We term this structural choke point the \textit{information compression penalty}. Regardless of the reasoning engine's capacity, its visual pipeline remains a low-bandwidth, lossy conduit.

\subsection{Formulating the Divergence Law}
We hypothesise the Divergence Law of Multimodal Scaling to analyse how this penalty manifests as the models grow. As model parameters scale by orders of magnitude, macroscopic benchmark accuracy ($S_{Full}$) will increase. This rising metric is conventionally taken as evidence of progress. However, from the lens of the Modality Translation Protocol, this overstates true multimodal capability.

\begin{figure}[h]
    \centering
    \includegraphics[width=\columnwidth]{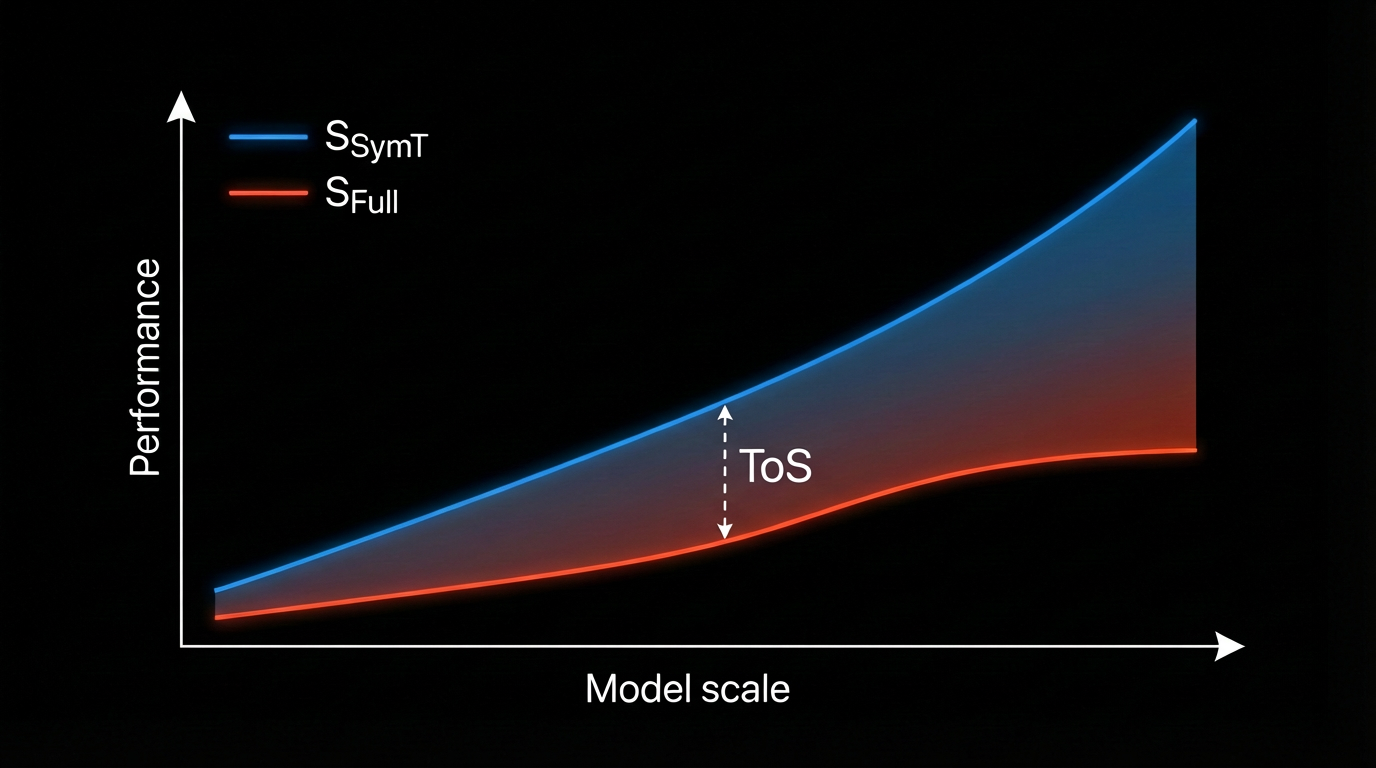}
    \caption{The Divergence Law of Multimodal Scaling. As Model Scale in Parameters (x-axis) increases, the LLM's true reasoning capability ($S_{SymT}$, logarithmic curve) rises sharply, while the macroscopic benchmark performance ($S_{Full}$) scales more shallowly. The expanding shaded region between the two curves represents a growing Toll of Seeing ($ToS$).}
    \label{fig:scaling_law}
\end{figure}

Because the visual projection bottleneck cannot scale its representational bandwidth proportionally to the LLM's cognitive leap, the model's true symbolic reasoning ceiling ($S_{SymT}$) rises at a faster rate compared to $S_{Full}$. Consequently, we project that the Toll of Seeing will increase proportionally with the model scale, as shown in Figure \ref{fig:scaling_law}.

\subsection{The Illusion of Capability}
The implications of this Divergence Law for world models are significant. If $ToS$ widens as scale increases, it implies that the relative cost of the visual modality grows with architecture scale. 

Scaling may not resolve multimodal reasoning failures; instead, it can amplify the LLM's ability to exploit text priors, masking the underlying visual bottleneck. As the LLM's text-only capability grows, the incentive to bypass a weaker visual encoder via language priors grows with it. Continued scaling within the current monolithic paradigm is therefore unlikely to yield faithful multimodal synthesis on its own, and may instead amplify reliance on language priors.

\section{A Roadmap for KDD: From a Diagnostic Constraint to SSC-Guided Architectures}
The introduction of the Semantic Sufficiency Criterion (SSC) represents a paradigm shift that uniquely targets the core mandate of the KDD community. Historically, the field has relied on \textit{data scavenging}, i.e., scraping massive, noisy repositories of loosely correlated image-text pairs or even the manually intensive image-text pair creation, and relying on the false hope that blind, billion-parameter next token prediction will force emergent alignment. 

The Divergence Law of Multimodal Scaling suggests that passive scaling alone is insufficient. However, this does not mean we must abandon the monolithic paradigm (Vision Encoder-Projector-LLM). Instead, we propose that the KDD community elevate the SSC from a passive diagnostic constraint into an \textit{active architectural blueprint and training objective}. We persuade the community to pioneer four directions to support this new paradigm, aiming to improve compositional visual reasoning within monolithic VLMs:

\subsection{Semantic Equivalence Engineering (SEE)}
To operationalise SSC at an industrial scale, we cannot train on raw, unstructured data. We must architect datasets based on strict mathematical isomorphism. Future KDD algorithms should be designed to generate isomorphic multimodal tuples $(V, V_{label}, T, T_{img})$ where the mutual information across modalities is entrusted to be equivalent. Utilising oracle-driven symbolic extraction, the community should engineer large-scale datasets that provide the strict $S_{SymT}$ baselines required for SSC-guided training.

\subsection{SSC as an Objective Function}
The Systems for Data Science community should move beyond static benchmark leaderboards and cross-entropy loss. We envision the development of \textit{Information-Theoretic Alignment Optimisation}. KDD researchers can formulate the Toll ($ToS$), Curse ($CoS$), and Fallacy ($FoS$) of Seeing into dynamic regularisation penalties during the VLM pre-training and alignment phases. By penalising the model when $S_{Full}$ diverges from $S_{SymT}$ or $S_{SymV}$, the loss landscape can be shaped to discourage over-reliance on text priors and encourage faithful visual feature extraction and alignment. 

\subsection{Architecting the Faithful Monolithic Paradigm}
By utilising $FoS$ as a localised gradient signal, we can address the information compression penalty inherent in monolithic VLMs. For example, when the Negative Collapse Mode ($FoS < 0$) is detected during training, it signals integration failure. KDD systems can use this real-time feedback to dynamically route, expand, or regularise the bandwidth of the projection head. By using the SSC to guide architectural decisions during training, compositional visual reasoning may be achievable within the monolithic paradigm, aligning the continuous visual manifold with the discrete token space.

\subsection{Dynamic SSC Auditing Architectures}
We envision the development of Dynamic SSC Auditing Engines--autonomous, data modulating systems deployed in production. These systems will continuously perturb incoming visual data streams, translating them into $S_{SymT}$ and $S_{SymV}$ on the fly to monitor a deployed VLM’s Expense of Seeing in real time. This ensures that a model’s knowledge extraction remains faithful under distributional shifts in a high-stakes inference scenario.

\section{If Successful: Implications for Faithful World Models}
If the KDD community embraces the ``Expense of Seeing'' framework, it could meaningfully shift practice in four areas:

\subsection{Redefining State-of-the-Art Benchmarking}
We will witness a clearer separation between models that genuinely use visual input and those that do not. By operationalising the SSC, the KDD community will systematically identify and deprioritise models that disguise textual statistical guessing as visual reasoning. We will no longer celebrate a model that achieves $80\%$ visual accuracy if its symbolic text ceiling ($S_{SymT}$) is $95\%$. The resulting $15\%$ \textit{Toll of Seeing} will be recognised as a representation bottleneck, rather than celebrated as a benchmark win. 

\subsection{Achieving True CVR Capability via Monolithic Paradigm}
A successful outcome would be reliable compositional visual reasoning within the monolithic paradigm. By enforcing SSC constraints during the training loop, monolithic architectures may be designed such that visual and textual representations share a more symmetric latent space. This would suggest the monolithic paradigm is not inherently flawed; rather, the historical reliance on metrics like accuracy and unconstrained, ablative training data was blinding it.

\subsection{Implications for Regulating Multimodal AI}
As we push toward foundational world models entrusted with human lives (e.g., autonomous medical diagnostics, algorithmic trading, and response management systems), trust cannot be heuristic; it must be provable. The SSC could potentially inform regulatory standards for Trustworthy Multimodal AI. One could imagine future regulatory frameworks requiring developers to demonstrate that $\max(ToS, CoS, |FoS|) \approx 0$, indicating that the model extracts truth faithfully without asymmetric modality bias.

\subsection{Rethinking Scaling Priorities}
This framework offers an alternative to purely compute-driven scaling that currently dominates the AI industry. We will shift the current infrastructure race away from merely maximising parameter counts and next token prediction. Success means the community broadens how it measures multimodal capability: true capability will no longer be measured by how much data a model can ingest, but by its modality symmetry. 

By transitioning from asking \textit{``Are models working?''} to establishing a rigorous, trustworthy diagnostic constraint that directly guides architectural training, the KDD community will lay a foundation for evaluating multimodal world models more rigorously.

\begin{acks}
To Dikshant Kukreja, for the framing of $ToS$.
\end{acks}

\bibliographystyle{ACM-Reference-Format}
\bibliography{references}

\end{document}